\documentclass{article}

\usepackage{microtype}
\usepackage{graphicx}
\usepackage{subfigure}
\usepackage{booktabs} 

\usepackage{hyperref}

\usepackage[accepted]{icml2023}

\usepackage{amsmath}
\usepackage{amssymb}
\usepackage{mathtools}
\usepackage{amsthm}

\usepackage[capitalize,noabbrev]{cleveref}

\theoremstyle{plain}

\theoremstyle{definition}

\theoremstyle{remark}

\usepackage[textsize=tiny]{todonotes}
\usepackage{hyperref}

\icmltitlerunning{Measuring the Success of Diffusion Models at Imitating Human Artists}

\begin{document}

\twocolumn[
\icmltitle{Measuring the Success of Diffusion Models at Imitating Human Artists}



\icmlsetsymbol{equal}{*}

\begin{icmlauthorlist}
\icmlauthor{Stephen Casper}{equal,MIT}
\icmlauthor{Zifan Guo}{equal,MIT}

\icmlauthor{Shreya Mogulothu}{MIT}
\icmlauthor{Zachary Marinov}{MIT}
\icmlauthor{Chinmay Deshpande}{Harvard}
\icmlauthor{Rui-Jie Yew}{MIT,Brown}
\icmlauthor{Zheng Dai}{MIT}

\icmlauthor{Dylan Hadfield-Menell}{MIT}
\end{icmlauthorlist}

\icmlaffiliation{MIT}{MIT}
\icmlaffiliation{Harvard}{Harvard University}
\icmlaffiliation{Brown}{Brown University}

\icmlcorrespondingauthor{Stephen Casper}{scasper@mit.edu}

\icmlkeywords{Machine Learning, ICML}

\vskip 0.3in
]



\printAffiliationsAndNotice{\icmlEqualContribution} 


\section*{Overview}

Modern diffusion models have set the state-of-the-art in AI image generation. 
Their success is due, in part, to training on Internet-scale data which often includes copyrighted work. This prompts questions about the extent to which these models learn from, imitate, or copy the work of human artists. 

This work suggests that questions involving copyright liability should factor in a model's \emph{capacity} to imitate an artist. 
Tying copyright liability to the capabilities of the model may be useful given the evolving ecosystem of generative models. 
Specifically, much of the legal analysis of copyright and generative systems focuses on the use of protected data for training \cite{sag2018new, lemley2020fair}. 
However, generative systems are often the result of multiple training processes. As a result, the connections between data, training, and the system are often obscured. 

\begin{figure}[t!]
    \centering
    \includegraphics[width=0.7\linewidth]{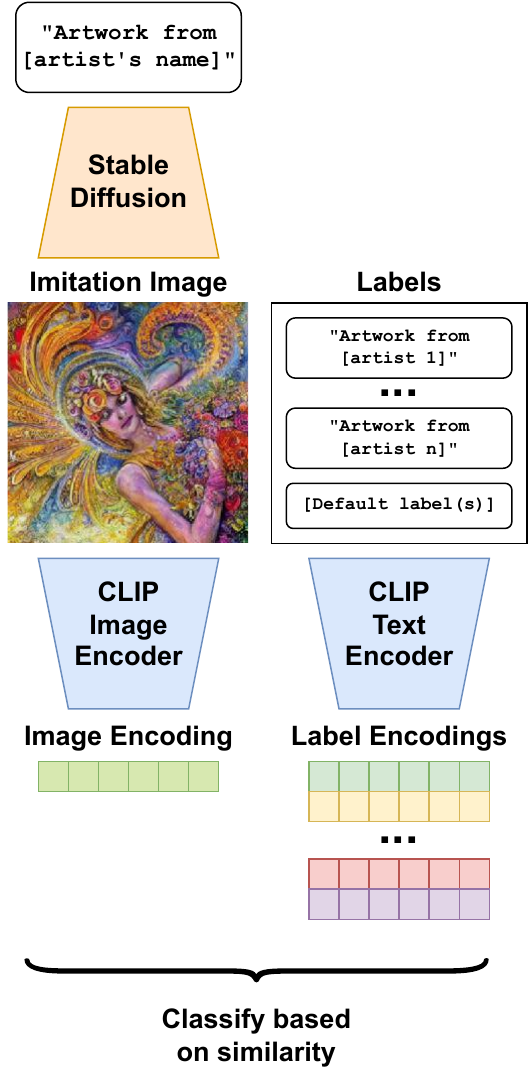}
    \vspace{-0.1in}
    \caption{\textbf{Identifying human artists from Stable Diffusion Imitations.} For each artist, we generate an imitation image from Stable Diffusion with the prompt ``Artwork from $<$ artist name $>$.'' Next, we encode the image with a CLIP image encoder \cite{radford2021learning}. We also encode labels corresponding to $n$ total artists plus one or more `default' labels with a CLIP text encoder. Finally, we classify the image among all labels using a geometric similarity measure between the encodings. If the label reliably corresponds to the correct artist, we consider the model to have the capability to imitate that artist.\vspace{-30pt}}
    \label{fig:method}
\end{figure}

\begin{figure*}[h!]
    \centering
    \includegraphics[width=0.93\linewidth]{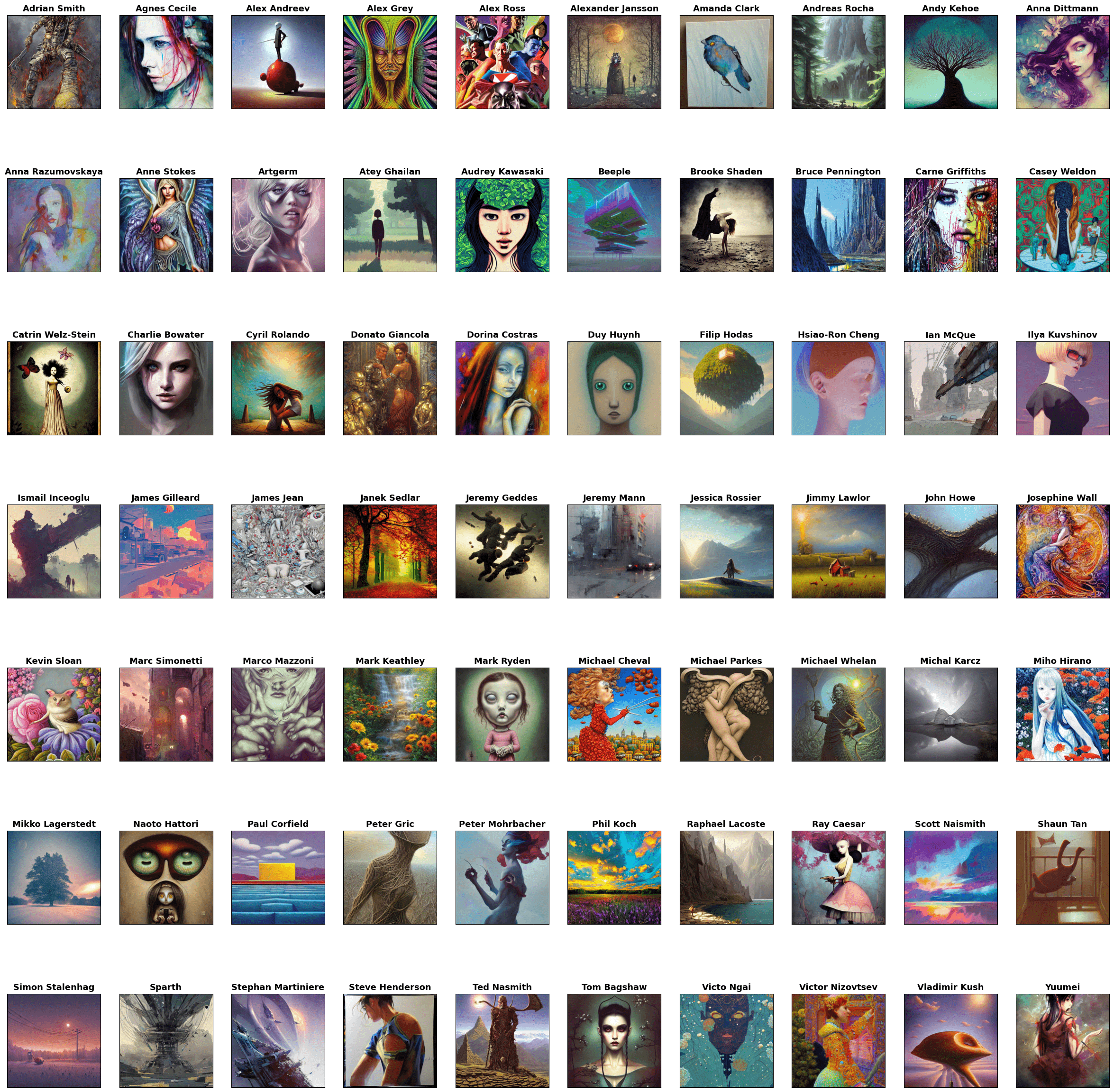}
    \caption{\textbf{Example images generated by Stable Diffusion from prompts of the form ``Artwork from $<$artist’s name$>$''.} Using the method depicted in Figure \ref{fig:method}, we show that the artists used in the prompts can often be classified from these imitations of their work.} 
    \label{fig:70_artists}
\end{figure*}

In our approach, we consider simple image classification techniques to measure a model's ability to imitate specific artists. Specifically, we use Contrastive Language-Image Pretrained (CLIP)~\citep{radford2021learning} encoders to classify images in a zero-shot fashion. 
Our process first prompts a model to imitate a specific artist. Then, we test whether CLIP can be used to reclassify the artist (or the artist's work) from the imitation. If these tests match the imitation back to the original artist, this suggests the model can imitate that artist's expression.

Our approach is simple and quantitative. Furthermore, it uses standard techniques and does not require additional training. We demonstrate our approach with an audit of Stable Diffusion's~\citep{rombach2022high} capacity to imitate 70 professional digital artists with copyrighted work online. When Stable Diffusion is prompted to imitate an artist from this set, we find that the artist can be identified from the imitation with an average accuracy of 81.0\%. Finally, we also show that a sample of the artist's work can be matched to these imitation images with a high degree of statistical reliability. Overall, these results suggest that Stable Diffusion is broadly successful at imitating individual human artists. Code is available \href{https://colab.research.google.com/drive/1ScHo9uMdUgId0DlSr4W4RgnMD44dLiku?usp=sharing}{here}.


\section{Background}

\textbf{Contrastive Language-Image Pretraining (CLIP):} CLIP \cite{radford2021learning} is a technique for training AI systems that encode images and text into fixed-length vector representations. 
CLIP image and text encoders are trained to produce similar encodings of image/caption pairs and dissimilar encodings of image/caption non-pairs. 
The more geometrically distant two encodings of images or captions are, the less related they are according to the encoder, and vice versa. 
Using this principle, \citet{radford2021learning} introduced a method to classify an image among a set of labels based on the distances between encodings. We use this method in our proposed test.

\textbf{Diffusion Models:} Diffusion models \cite{sohl2015deep} such as Stable Diffusion \citep{rombach2022high} and Midjourney \citep{midjourney}, are capable of generating images from arbitrary, user-specified prompts. 
Their success has largely been due to training on large amounts of text/image data, often including copyrighted works \cite{schuhmann2021laion}.
Modern image-generation diffusion models are trained using CLIP-style encoders. 
When given an encoding of a caption, a diffusion model is trained to generate an image corresponding to the caption \cite{ramesh2022hierarchical}. 
Accordingly, a diffusion model that generates images from these embeddings is trained to be the inverse of a CLIP image encoder.

\textbf{Legal Motivation:} In the United States, \citet{newtonvdiamond} established that copyright infringement ``is measured by considering the qualitative and quantitative significance of the copied portion in relation to the plaintiff’s work as a whole''. However, the subjective nature of these determinations makes practical enforcement complicated.
\cite{balganesh2014judging, kaminski2017copyright, balagopalan2023judging}.
In evaluating copyright questions involving AI systems, legal analyses have focused on how copyrighted work is used in the system's training data~\cite{sag2018new, lemley2020fair}, but such a focus on training data does not connect liability to an AI system's ability to copy an artist. 
In contrast, we show how standard image classification techniques can be used to help determine how successful AI image generators are at imitating individual human artists. 
This approach is \textit{consistent}, \textit{quantitative}, and connected to the \emph{capabilities} of the resulting AI system. 
Our goal, however, is not to automate determinations of infringement but to demonstrate how tried and tested image classification techniques from machine learning can be used to analyze legal claims.


\begin{figure}[t]
    \centering
    \includegraphics[width=0.9\linewidth]{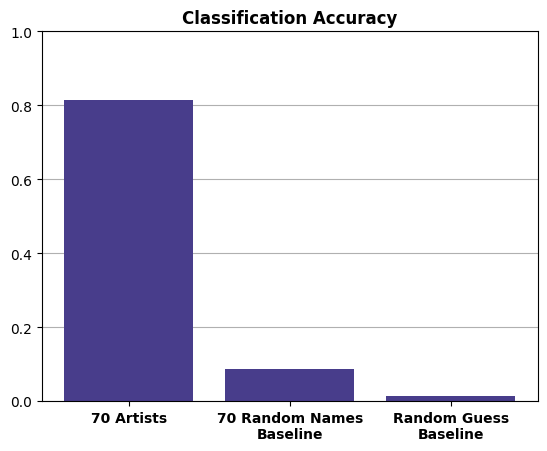}
    \vspace{-0.1in}
    \caption{\textbf{Results from human artists strongly outperform baselines.} Success rates for classifying artists from Stable Diffusion's attempts to imitate them. Professional artists can be classified from imitations over 81.0\% of the time on average. This compares to 8.6\% for a baseline in which we used random names instead of artists' names and 1.4\% for a random guess baseline.\vspace{-15pt}}
    \label{fig:barplot}
\end{figure}

\begin{figure*}[h]
    \centering
    \includegraphics[width=0.7\linewidth]{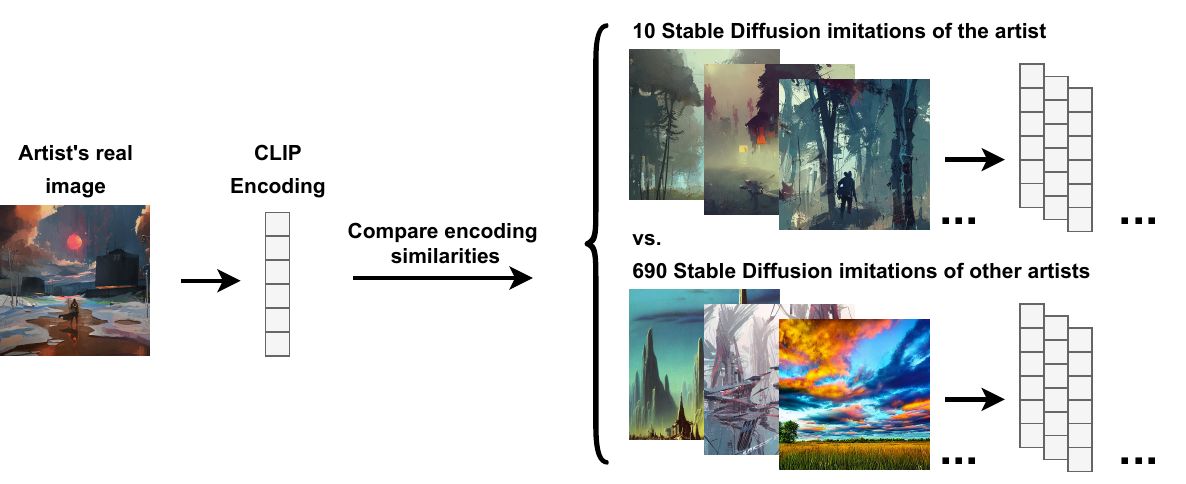}

    \vspace{0.1in}
    
    \hrule

    
    \includegraphics[width=\linewidth]{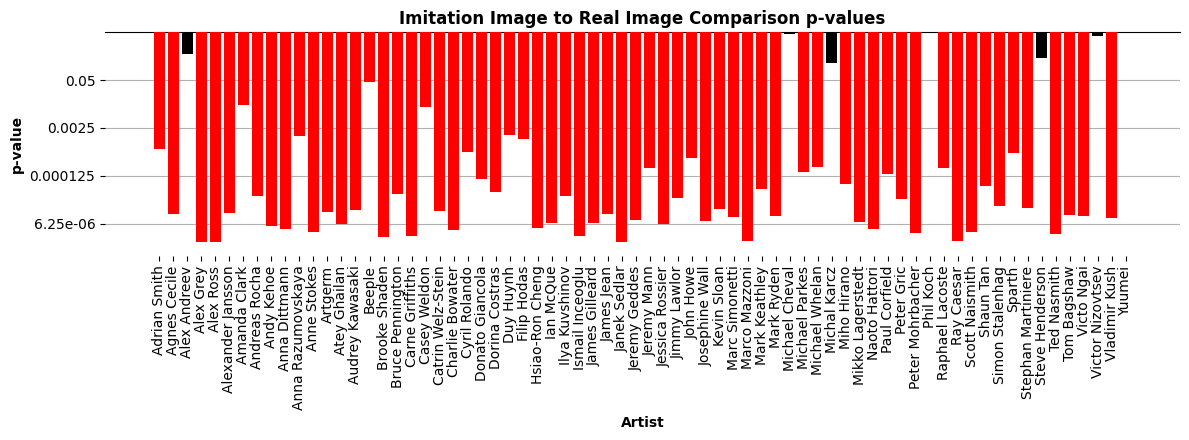}
    \vspace{-0.2in}
    \caption{\textbf{Matching human artwork to imitations from Stable Diffusion:} (Top) Method: We compared the encoding of one real image per artist to 10 imitations of that artist and 690 imitations of the other artists. Then we aggregated these results with a statistical rank sum test. (Bottom) Results: Artwork generated by Stable Diffusion with the prompt ``Artwork from $<$artist's name$>$'' is significantly more similar to real artwork by the artist in question than artwork generated to imitate other artists. Artists for which the experiment resulted in a (Bonferroni-corrected) p-value below 0.05 are highlighted in red. This occurred for 90\% (63/70) of the artists.  \vspace{-5pt}} 
    \label{fig:real_artists}
\end{figure*}

\section{Experiments}

We conduct two complementary experiments to evaluate Stable Diffusion's ability to imitate human artists. First, we classify human artists from imitations of their work, and second, we match real work from human artists to imitations. Both experiments suggest that Stable Diffusion is broadly successful at imitating human artists. 

\subsection{Identifying Artists from Imitations}

\textbf{Method:} We used CLIP encoders to classify artists from Stable Diffusion's imitations of them. We selected 70 artists from the LAION-aesthetics dataset \cite{schuhmann2021laion}, the dataset used to train Stable Diffusion. We selected these 70 as artists who may potentially be harmed by digital imitations using several criteria: each artist is alive, has a presence on digital art platforms (Instagram, DeviantArt, and ArtStation), publishes artwork or sells their artwork (e.g., prints or digital works), and has more than 100 images in the LAION dataset.

Figure \ref{fig:method} outlines our method. 
We prompted \href{https://huggingface.co/runwayml/stable-diffusion-v1-5}{Stable Diffusion (v1.5)} to generate images in the style of each artist, using prompts of the form ``Artwork from $<$artist’s name$>$''. 
Example images are in Figure \ref{fig:70_artists}.
We then used \href{https://huggingface.co/openai/clip-vit-base-patch32}{CLIP encoders} to classify each image among a set of 73 labels.
The 73 labels consisted of each of the 70 artist's prompts (``Artwork from $<$artist’s name$>$'') plus three default labels: ``Artwork'', ``Digital Artwork'', and ``Artwork from the public domain.'' 
These additional labels lend insight into how confident CLIP is that an image imitates a particular artist's style instead of some more generic style. 
We then classified each imitation image among these labels using the technique from \citet{radford2021learning}. 
CLIP-based classification produces a probability of an image matching each label, and we evaluate the model on the correctness of its most-likely prediction and confidence in the correct artists.

\textbf{Results:} We repeated the experiment with the 70 artists ten times to reduce the effect of random variation. On average, CLIP correctly classified 81.0\% of the generated images as works made by artists whose names were used to generate them. 
Over the ten trials, 69 of the 70 artists were correctly classified in a plurality of the ten trials.
Overall, these results suggest that Stable Diffusion has a broad-ranging ability to imitate the styles of individual artists. 
We compared these results to two baselines.
First, we implemented a random-name baseline by running the same experiment with 70 random names from a \href{https://randomwordgenerator.com/name.php}{random name generator}.
Since Stable Diffusion was not trained on artists with these names (unless a random name is coincidentally the same as some artist's), this experiment serves as a proxy for how Stable Diffusion would handle artists not in its training data. 
In this case, only 6 names (8.6\%) were guessed correctly.
Second, a random guess would only result in a successful classification every 1 in 73 attempts (1.4\%) on average.
We visualize results from our main experiment alongside the controls in Figure \ref{fig:barplot}.

\textbf{Results are Robust to Different Sets of Artists:} To test whether our 70 artists were especially classifiable, we ran the original experiment but with a larger set of indiscriminately-selected artists and found similar results. We selected the 250 artists with the highest number of images in the LAION dataset and found that CLIP correctly classified 81.2\% of the images. 
This demonstrates that successful classification transcends a particular specific set of artists.

\subsection{Matching Artwork to Imitations}

\textbf{Method:} Our first experiment tested how easily artists could be identified from diffusion model imitations of them. 
To provide a complementary perspective, we also directly study the similarity of artists' digital works to Stable Diffusion's imitations of them. For each of the 70 artists, we retrieve the top result obtained by Google Image searching ``$<$artist's name$>$ art.''
As before, we then use Stable Diffusion to generate 10 images for each artist with the prompt ``Artwork from [artist's name].'' We then compare the real images and generated images. Distances are measured by first encoding images 
using the CLIP image encoder and calculating the cosine distance between encodings. 

\textbf{Results:} For each artist, we calculate whether real images from artists are more similar to imitations of that artist or other artists. The significance was calculated using a rank sum test with a Bonferroni correction factor of 70. Results are in Figure \ref{fig:real_artists}.
90\% (63/70) of the experiments produce $p$ values less than 0.05. This compares to an average of 22.8\% (16/70) for a control experiment using random artist assignments of real images. These results further support that Stable Diffusion is broadly successful at imitating artists.

\section{Conclusion}

We have demonstrated how AI image classification can help to measure the success of diffusion models imitating human artists. 
We argue that these methods can provide a practical way to tie questions about copyright liability to the \emph{capabilities} of a model instead of its training data alone. 
By matching imitation images to both artists' names and works, we find that Stable Diffusion is broadly successful at imitating human digital artists.
We hope that future work can use image classification to analyze legal claims and to test defenses against AI imitation of copyrighted work.

\section*{Acknowledgements}

We thank Taylor Lynn Curtis and Lennart Schulze for feedback.

\bibliography{bibliography}

\begin{thebibliography}{12}
\providecommand{\natexlab}[1]{#1}
\providecommand{\url}[1]{\texttt{#1}}
\expandafter\ifx\csname urlstyle\endcsname\relax
  \providecommand{\doi}[1]{doi: #1}\else
  \providecommand{\doi}{doi: \begingroup \urlstyle{rm}\Url}\fi

\bibitem[Balagopalan et~al.(2023)Balagopalan, Madras, Yang, Hadfield-Menell,
  Hadfield, and Ghassemi]{balagopalan2023judging}
Balagopalan, A., Madras, D., Yang, D.~H., Hadfield-Menell, D., Hadfield, G.~K.,
  and Ghassemi, M.
\newblock Judging facts, judging norms: Training machine learning models to
  judge humans requires a modified approach to labeling data.
\newblock \emph{Science Advances}, 9\penalty0 (19):\penalty0 eabq0701, 2023.

\bibitem[Balganesh et~al.(2014)Balganesh, Manta, and
  Wilkinson-Ryan]{balganesh2014judging}
Balganesh, S., Manta, I.~D., and Wilkinson-Ryan, T.
\newblock Judging similarity.
\newblock \emph{Iowa L. Rev.}, 100:\penalty0 267, 2014.

\bibitem[Kaminski \& Rub(2017)Kaminski and Rub]{kaminski2017copyright}
Kaminski, M.~E. and Rub, G.~A.
\newblock Copyright's framing problem.
\newblock \emph{UCLA L. Rev.}, 64:\penalty0 1102, 2017.

\bibitem[Lemley \& Casey(2020)Lemley and Casey]{lemley2020fair}
Lemley, M.~A. and Casey, B.
\newblock Fair learning.
\newblock \emph{Tex. L. Rev.}, 99:\penalty0 743, 2020.

\bibitem[Midjourney(2022)]{midjourney}
Midjourney.
\newblock Midjourney, 2022.
\newblock URL \url{https://www.midjourney.com/}.

\bibitem[Radford et~al.(2021)Radford, Kim, Hallacy, Ramesh, Goh, Agarwal,
  Sastry, Askell, Mishkin, Clark, et~al.]{radford2021learning}
Radford, A., Kim, J.~W., Hallacy, C., Ramesh, A., Goh, G., Agarwal, S., Sastry,
  G., Askell, A., Mishkin, P., Clark, J., et~al.
\newblock Learning transferable visual models from natural language
  supervision.
\newblock In \emph{International conference on machine learning}, pp.\
  8748--8763. PMLR, 2021.

\bibitem[Ramesh et~al.(2022)Ramesh, Dhariwal, Nichol, Chu, and
  Chen]{ramesh2022hierarchical}
Ramesh, A., Dhariwal, P., Nichol, A., Chu, C., and Chen, M.
\newblock Hierarchical text-conditional image generation with clip latents.
\newblock \emph{arXiv preprint arXiv:2204.06125}, 2022.

\bibitem[Rombach et~al.(2022)Rombach, Blattmann, Lorenz, Esser, and
  Ommer]{rombach2022high}
Rombach, R., Blattmann, A., Lorenz, D., Esser, P., and Ommer, B.
\newblock High-resolution image synthesis with latent diffusion models.
\newblock In \emph{Proceedings of the IEEE/CVF Conference on Computer Vision
  and Pattern Recognition}, pp.\  10684--10695, 2022.

\bibitem[Sag(2018)]{sag2018new}
Sag, M.
\newblock The new legal landscape for text mining and machine learning.
\newblock \emph{J. Copyright Soc'y USA}, 66:\penalty0 291, 2018.

\bibitem[Schuhmann et~al.(2021)Schuhmann, Vencu, Beaumont, Kaczmarczyk, Mullis,
  Katta, Coombes, Jitsev, and Komatsuzaki]{schuhmann2021laion}
Schuhmann, C., Vencu, R., Beaumont, R., Kaczmarczyk, R., Mullis, C., Katta, A.,
  Coombes, T., Jitsev, J., and Komatsuzaki, A.
\newblock Laion-400m: Open dataset of clip-filtered 400 million image-text
  pairs.
\newblock \emph{arXiv preprint arXiv:2111.02114}, 2021.

\bibitem[Sohl-Dickstein et~al.(2015)Sohl-Dickstein, Weiss, Maheswaranathan, and
  Ganguli]{sohl2015deep}
Sohl-Dickstein, J., Weiss, E., Maheswaranathan, N., and Ganguli, S.
\newblock Deep unsupervised learning using nonequilibrium thermodynamics.
\newblock In \emph{International Conference on Machine Learning}, pp.\
  2256--2265. PMLR, 2015.

\bibitem[\textup{\textit{Newton v. Diamond}, 388 F.3d 1189, 1195 (9th Cir.
  2004)}()]{newtonvdiamond}
\textup{\textit{Newton v. Diamond}, 388 F.3d 1189, 1195 (9th Cir. 2004)}.

\end{thebibliography}
\bibliographystyle{icml2023}



\end{document}